\documentclass{ieeeaccess}
\usepackage{cite}
\usepackage{amsmath,amssymb,amsfonts}
\usepackage{bm}
\usepackage{algorithmic}
\usepackage{graphicx}
\usepackage{textcomp}
\usepackage{algorithm,algorithmic}
\usepackage[acronym]{glossaries}
\usepackage{color, soul}

\def\BibTeX{{\rm B\kern-.05em{\sc i\kern-.025em b}\kern-.08em
    T\kern-.1667em\lower.7ex\hbox{E}\kern-.125emX}}

\newacronym{mv}{MV}{mitral valve}
\newacronym{mvp}{MVP}{mitral valve prolapse}
\newacronym{mr}{MR}{mitral regurgitation}
\newacronym{3dtee}{3DTEE}{three-dimensional TEE}

\begin{document}
\history{Date of publication xxxx 00, 0000, date of current version xxxx 00, 0000.}
\doi{10.1109/ACCESS.2023.0322000}

\title{A Deep Learning-Based Fully Automated Pipeline for Regurgitant Mitral Valve Anatomy Analysis from 3D Echocardiography}
\author{\uppercase{Riccardo Munafò}\authorrefmark{1}, %\IEEEmembership{Fellow, IEEE},
Simone Saitta\authorrefmark{1}, Giacomo Ingallina\authorrefmark{2}, Paolo Denti\authorrefmark{3}, Francesco Maisano\authorrefmark{3}, Eustachio Agricola \authorrefmark{2,4}, Alberto Redaelli\authorrefmark{1} and Emiliano Votta\authorrefmark{1}}

\address[1]{Department of Electronics, Information and Bioengineering, Politecnico di Milano, 20133 Milan, Italy (e-mail: riccardo.munafo@polimi.it; simone.saitta@polimi.it; alberto.redaelli@polimi.it; emiliano.votta@polimi.it).}
\address[2]{Unit of Cardiovascular Imaging, IRCCS San Raffaele Hospital, 20132 Milan, Italy (e-mail: ingallina.giacomo@hsr.it, eustachio.agricola@hsr.it).}
\address[3]{Cardiac Surgery Department, IRCCS San Raffaele Hospital, 20132 Milan, Italy (e-mail: paolo.denti@hsr.it; francesco.maisano@hsr.it).}
\address[4]{Vita-Salute San Raffaele University, 20132 Milan, Italy (e-mail: eustachio.agricola@hsr.it).}

\tfootnote{This work has received funding from the European Union’s Horizon 2020research and innovation programme under grant agreement No. 101017140, the ARTERY project.}

\markboth
{Munafò \headeretal: A Deep Learning-Based Fully Automated Pipeline for Regurgitant Mitral Valve Anatomy Analysis from 3D Echocardiography}
{Munafò \headeretal: A Deep Learning-Based Fully Automated Pipeline for Regurgitant Mitral Valve Anatomy Analysis from 3D Echocardiography}

\corresp{Corresponding author: Riccardo Munafò (e-mail: riccardo.munafo@polimi.it).}

\begin{abstract}
Three-dimensional transesophageal echocardiography (3DTEE) is the recommended imaging technique for the assessment of mitral valve (MV) morphology and lesions in case of mitral regurgitation (MR) requiring surgical or transcatheter repair. Such assessment is key to thorough intervention planning and to intraprocedural guidance. However, it requires segmentation from 3DTEE images, which is time-consuming, operator-dependent, and often merely qualitative. In the present work, a novel workflow to quantify the patient-specific MV geometry from 3DTEE is proposed. The developed approach relies on a 3D multi-decoder residual convolutional neural network (CNN) with a U-Net architecture for multi-class segmentation of MV annulus and leaflets. The CNN was trained and tested on a dataset comprising 55 3DTEE examinations of MR-affected patients. After training, the CNN is embedded into a fully automatic, and hence fully repeatable, pipeline that refines the predicted segmentation, detects MV anatomical landmarks and quantifies MV morphology. The trained 3D CNN achieves an average Dice score of 0.82 $\pm$ 0.06, mean surface distance of 0.43 $\pm$ 0.14 mm and 95\% Hausdorff Distance (HD) of 3.57 $\pm$ 1.56 mm before segmentation refinement, outperforming a state-of-the-art baseline residual U-Net architecture, and provides an unprecedented multi-class segmentation of the annulus, anterior and posterior leaflet. The automatic 3D linear morphological measurements of the annulus and leaflets, specifically diameters and lengths, exhibit differences of less than 1.45 mm when compared to ground truth values. These measurements also demonstrate strong overall agreement with analyses conducted by semi-automated commercial software. The whole process requires minimal user interaction and requires approximately 15 seconds.
\end{abstract}


\titlepgskip=-21pt

\maketitle

\section{Introduction}
\label{sec:introduction}
The \gls{mv} is a morphologically and functionally complex structure located between the left atrium and the left ventricle. It consists of two leaflets, whose bases are inserted on a saddle-shaped annulus that silhouettes the valve orifice, and whose free margins are connected to the myocardium of the left ventricle by a set of collagenous \textit{chordae tendineae}. In physiologic conditions, during ventricular diastole the \gls{mv} allows for ventricular filling and plays a role in the generation of 3D toroidal vortices that contribute to optimizing the energetics of diastolic intraventricular blood flow \cite{Martínez_2014}. During ventricular systole, the increase in transvalvular pressure difference drive \gls{mv} leaflet motion towards the atrium; the synergistic function of all \gls{mv} substructures allows for leaflet coaptation and \gls{mv} continence, thus preventing blood backflow into the atrium. Also, the physiological shape of \gls{mv} leaflets in their closed configuration has been suggested to contribute to reducing \gls{mv} leaflet tissue stresses \cite{Stevanella_2011} and to optimizing blood flow energetics during systolic ejection through the aortic valve and into the aorta \cite{Kvitting_2010, Dimasi_2012}.\\
A dysfunction of any \gls{mv} substructure typically leads to a loss of \gls{mv} continence, i.e., to \gls{mr}. \gls{mr} induces ventricular volume overload \cite{Gaasch_2008} and increases atrial pressure \cite{ORourke_1984}. When \gls{mr} is severe and remains untreated, volume overload induces the enlargement of the ventricular chamber, and hence excessive pre-load of the myocardial fibers that is detrimental to their efficient contraction, as explained by the Starling law \cite{Gaasch_2008}. On the long run, this anomaly can lead to heart failure \cite{Nishimura_2018, McDonagh_2021}. In fact, in untreated patients affected by severe \gls{mr} the mortality rate at 5 years is 50\%, and 90\% of those who survive are  hospitalized because of heart failure \cite{Goel_2014}. On the other hand, increased atrial pressure  often leads to pulmonary hypertension \cite{Opotowsky_2020}. 
Of note, \gls{mr} is the second most prevalent heart valve disease, and its most common cause is \gls{mvp}, i.e., the invasion of the atrial volume by a portion of \gls{mv} leaflets with loss of coaptation. \gls{mvp} alone affects 2-3\% of the general population \cite{de2016role}.
The preferred treatment for \gls{mr} is \gls{mv} repair, owing to its better short-term results and to significantly better long-term survival rates in the elderly population, as compared to \gls{mv} replacement \cite{DiTommaso_2021}. Moreover, percutaneous approaches for \gls{mv} repair have emerged as a valid alternative to open-chest \gls{mv} surgery. Among the available approaches, transcatheter edge-to-edge repair (TEER) is the most widespread one and is supported by the strongest evidence \cite{hearts2010007}. \\ 
According to the guidelines of the European Society of Cardiology (ESC) \cite{Vahanian_2021} and the American Heart Association (AHA) \cite{Otto_2020}, transthoracic echocardiography (TTE) is the standard imaging modality to diagnose and quantify \gls{mr}, while transesophageal echocardiography (TEE) is used for morphologic assessment of pathologic MV and to guide \gls{mv} procedures. Namely, \gls{3dtee} allows for a 3D view of the \gls{mv} and its substructures. Thus, it is widely used to identify and characterize anatomical defects of regurgitant \gls{mv}, to drive final decision and timing of surgical treatment, and to intraoperatively support TEER procedures by guiding the catheter into the left atrium and ensuring proper placement of the device to be implanted. However, accurate assessment of \gls{mv} morphology by \gls{3dtee} typically relies on manual or semi-automatic segmentation, making it time-consuming and affected by intra-operator and inter-operator variability, hence hampering the reliability of the resulting anatomical measurements. Precise \gls{mv} segmentation and identification of \gls{mv} annulus and leaflets from \gls{3dtee} would facilitate accurate and quantitative measurements of the regurgitant \gls{mv} anatomy to support \gls{mr} diagnosis and \gls{mv} surgical or transcatheter repair \cite{Zamorano_2011}.\\ 
In recent years, several works have introduced semi- \cite{Burlina_2010, Schneider_2011, Pouch_2017} or fully automatic \cite{Ionasec_2011, Andreassen_2019, Zhang_2020, Carnahan_2021, Chen_2023} methods to detect and segment \gls{mv} structures. Many early methods proposed in the last decade are based on level set \cite{Burlina_2010} or graph cut method \cite{Schneider_2011}, which require human-in-the-loop-interactions to work properly and reconstruct mitral leaflets as an inner surface lacking the preservation of leaflet thickness details. In \cite{Ionasec_2011}, a 3D geometrical model of \gls{mv} was constructed by fitting key points identified by leveraging machine learning algorithms and an extensive collection of labelled images. In \cite{Pouch_2017}, a semi-automatic method to segment mitral leaflets was described using multi-atlas joint label fusion and a deformable model template. However, abnormal morphologies typical of pathological \gls{mv} anatomy are often not accurately detected using a model fitting or atlas-based approach. In the last few years, methods based on deep learning, and in particular convolutional neural networks (CNNs), have become increasingly popular for \gls{mv} segmentation. In \cite{Andreassen_2019}, the authors proposed a method that leverages 2D CNN-based predictions obtained on individual image slices extracted from a \gls{3dtee} volume to generate a 3D annulus model through an iterative post-processing algorithm. Similarly, Zhang et al. \cite{Zhang_2020} applied a deep reinforcement learning algorithm and a 2D CNN for annulus landmark detection in combination with a spline fitting algorithm to infer the 3D mitral annulus from \gls{3dtee} images. In \cite{Carnahan_2021}, Carnahan et al. proposed a fully automatic method adopting a 3D CNN fed by \gls{3dtee} images for segmentation of \gls{mv} leaflets in diastole, demonstrating the feasibility of CNNs-based segmentation of \gls{3dtee} images. In a recent work, Chen et al. \cite{Chen_2023} developed an innovative pre-training strategy aiming to classify the diastolic and the systolic states of the \gls{mv}. This classification allowed to initialize the parameters of a 3D CNN for \gls{mv} segmentation in the entire cardiac cycle. 
Nevertheless, in all of these studies which leverage deep learning methods, \gls{mv} segmentation was formulated as a single-class semantic segmentation problem, detecting either the whole \gls{mv} apparatus \cite{Carnahan_2021, Chen_2023}, or simply the annulus \cite{Andreassen_2019, Zhang_2020}; hence, a solution to distinguish among the different substructures of the \gls{mv} is still missing. In this context, the proposed work aimed to improve the workflow for the extraction of complex patient-specific \gls{mv} geometric features from \gls{3dtee} by discerning among the annulus, anterior and posterior leaflets.\\
The system proposed in this study is a deep learning-based fully automatic pipeline for \gls{mv} substructure segmentation and anatomical characterization. A novel multi-decoder 3D CNN was trained to separately segment the mitral annulus, the anterior leaflet and the posterior leaflet from \gls{3dtee} acquisitions, under the hypothesis that the use of a multi-decoder architecture may improve performance in distinguishing spatially continuous structures within ultrasound images. The results of the automated segmentation automatically feed a processing module devoted to quantifying \gls{mv} annulus and leaflets morphology.\\
The multi-class segmentation is aimed to obtain a more comprehensive evaluation of \gls{mv} anatomy and defects as compared to previously proposed automated solutions. In particular, segmenting the two leaflets separately would enable the recostruction of the coaptation line, thus allowing for more reliable and automated \gls{mr} pre-procedural analysis and intraprocedural support. 
 
\section{Methods}
\label{sec:Methods}
\subsection{Data Collection}
Intraprocedural examinations were collected from 55 candidates to \gls{mv} TEER by means of the MitraClip system (Abbott) from IRCCS San Raffaele Hospital (Milan, Italy). Acquisitions were performed using a Philips EPIQ CVx scanner with an X8-2T transducer (Philips, Andover, MA, USA). Image data were anonymized and exported in Cartesian format using QLab software (Philips, Andover, MA, USA). The study was approved by the local ethics committee.\\
Midesophageal 4DTEE acquisitions zoomed on the \gls{mv} were selected from the collected examinations to construct the dataset. For some patients, multiple acquisitions were available and were included in the dataset. Acquisitions with high shadowing and those that did not entirely capture the \gls{mv} in the field of view of the ultrasound beam were excluded. The final dataset counted 115 \gls{3dtee} volumetric and time-dependent data with a mean voxel spacing of 0.37 $\times$ 0.55 $\times$ 0.24 mm\textsuperscript{3} and a mean frame rate of 22 per cardiac cycle. 
For each sequence, the end-systolic frame, identified as the time frame just before aortic valve closure, was selected.\\

\subsection{Manual Annotation}
\label{subsec: manual_annotation}
\begin{figure}[ht]
\centering
\centerline{\includegraphics[width=\columnwidth]{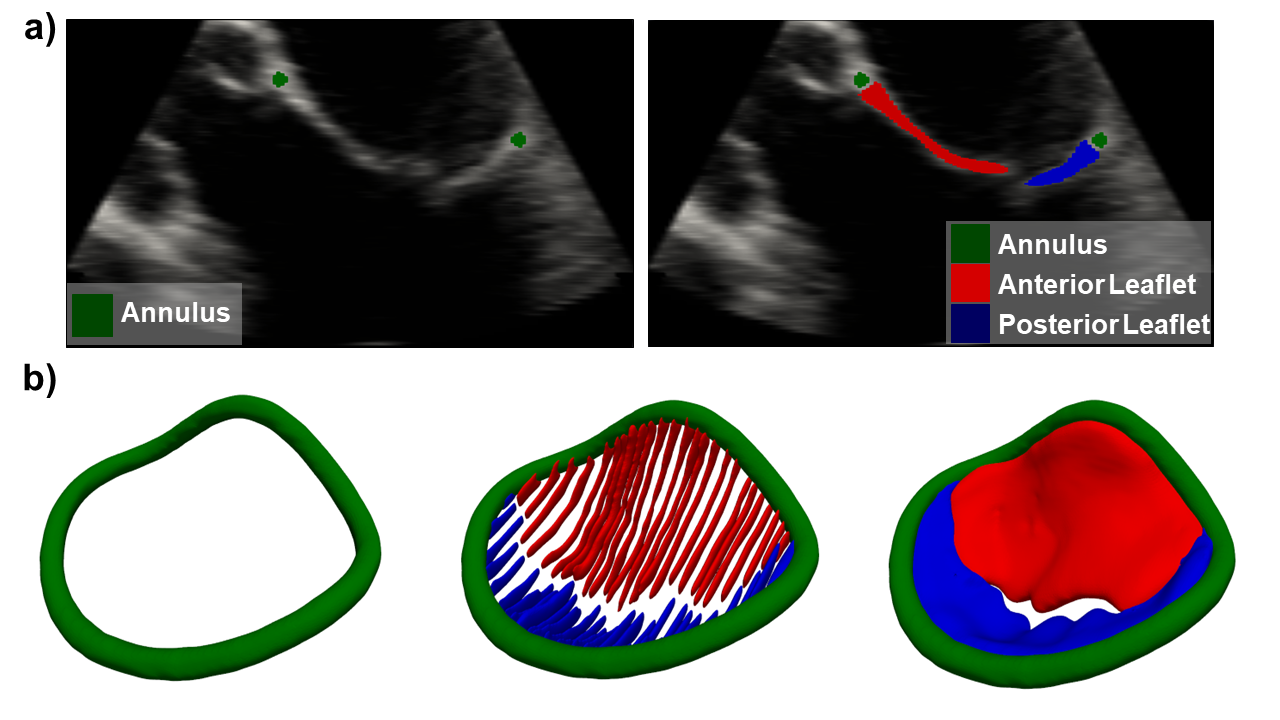}}
\caption{Intermediate and final result of the steps followed to manually annotate \gls{3dtee} images. (a) 3D rendering of the segmentation mask for each step, and (b) corresponding cross-sectional view overlapped on the \gls{3dtee} image.}
\label{fig:manual-segmentation}
\end{figure}

\gls{3dtee} data were manually annotated to generate the training and test sets for the deep learning models. Manual annotations were performed using 3D Slicer \cite{Kikinis_2014} by three independent and experienced operators, ensuring an almost balanced division of the dataset among them. Each operator followed the same protocol, consisting in the following steps:
\begin{itemize}
    \item selecting the end-systolic frame
    \item manually navigating the 3D data and identifying the standard 2-chamber view plane, where the LV long-axis is automatically generated
    \item automatically generating 12 long-axis view planes by rotating the 2D-chamber view plane around the LV long-axis so to span the [0;2$\pi$] range with a $\pi/12$ step
    \item manually tracing on each plane the two annular points, identified as the hinge points of the MV leaflets (Figure \ref{fig:manual-segmentation}a, left). This step resulted in 24 annular points, which were then fitted with a smooth curve using the Slicer-Heart software \cite{Lasso_2022} (Figure \ref{fig:manual-segmentation}b, left) with a contour radius of 1 mm 
    \item by manually adjusting the 2D-chamber view, obtaining a long-axis view plane that approximately runs through the annulus saddle horn (SH), i.e., the highest point on the anterior side of the annular profile, and the centroid of the annular profile
    \item manually navigating 3D data starting by shifting the obtained view plane along its normal, which approximately corresponded to the commissure-commissure (CC) axis of the annulus. In this process a variable number of mutually parallel image planes were generated so to i) span the whole MV orifice and ii) having consecutive planes separated by a distance equal to at least twice the minimum shift allowed for by 3D Slicer
    \item on each parallel long-axis view plane, manually segmenting the anterior and posterior leaflets, separated at the coaptation zone (Figure \ref{fig:manual-segmentation}a, right). This resulted in a non-contiguous segmentation of the leaflets on several long-axis slices (Figure \ref{fig:manual-segmentation}b, center), which were then interpolated (Figure \ref{fig:manual-segmentation}b, right) using a morphological contour interpolator filter \cite{Albu_2008}. 
\end{itemize}

\subsection{Analysis of Inter-operator Variability}
\label{subsec:interoperator variability}
To obtain an unbiased assessment of the inter-operator variability of MV segmentation using the protocol described above, a randomly selected subset of 10 \gls{3dtee} volumes belonging to 10 different patients from the dataset were manually annotated by three independent and experienced operators, named Op1, Op2 and Op3, respectively. All operators used 3D Slicer \cite{Kikinis_2014}, exploiting the 3D Slicer extension Slicer-Heart \cite{Lasso_2022}, and followed the protocol described in the subsection \ref{subsec: manual_annotation} to perform the annotation. The average time required by operators to perform the annotation was also considered in the analysis.

\subsection{Neural Network Architecture and Training}
\label{subsec:neural-network-architecture} 
The trained CNN was based on a modified 3D U-Net \cite{Ronneberger_2015} with encoding and decoding branches of 5 resolution levels each, defined using residual units as shown in \cite{Zhang_2018}. Each residual unit consists of 2 convolutional blocks and an identity mapping. Each convolutional block included a batch normalization layer, a parametric rectified linear unit (PReLU) activation layer \cite{He_2015} and a 3$\times$3 convolutional layer with stride of 2. In addition, the number of decoding branches was extended to three, one for each class detected (i.e., mitral annulus, anterior leaflet and posterior leaflet). This resulted in a multi-decoder version of the 3D residual U-Net.\\
For the training of the neural network, the collected data were randomly divided according to the number of examinations (55). Specifically, 90\% of the examinations (50) were allocated to the training set, while the remaining 10\% (5) were assigned to the test set. In the training set, all the available volumes for each examination with the corresponding ground truth (GT) annotations were considered, resulting in a total of 110 TEE volumes. To address the limited amount of available data and ensure a robust evaluation of the trained model, a 10-fold cross-validation was performed on the training set. The 3D CNN architecture was hence trained and evaluated ten times. Finally, the resulting 10 trained models were combined by stacking them together and taking the maximum predictions along each detected class. A data augmentation routine was adopted to increase the diversity of the data: transformations taken from the MONAI framework \cite{Cardoso_2022} were used including, Gaussian noise, cropping, flipping, rotation and elastic deformation; transformations were applied randomly to the dataset. The model was implemented using the Pytorch framework \cite{Pytorch} and it was trained on an NVIDIA A100 over 300 epochs with batch size of 4. A Novograd optimizer \cite{Ginsburg_2019} was used with initial learning rate of 0.001, reduced by a factor of 10 every 100 epochs. Intensity of the input images was normalized with values ranging between 0 and 1. A weighted-combination of Dice and Focal losses \cite{Yeung_2022} (with weight of 0.6 and 0.4, respectively) was computed at the output block of each decoder branch. The sum of the loss functions of the three decoders was used to train the neural network. Each model needed around 5 hours to be trained over 300 epochs and only the model with the best validation accuracy for each fold was saved and used for inference.

\subsection{Pipeline Implementation}
\label{subsec:pipeline-implementation}

\begin{figure}[ht]
 \centering
 \centerline{\includegraphics[width=\columnwidth]{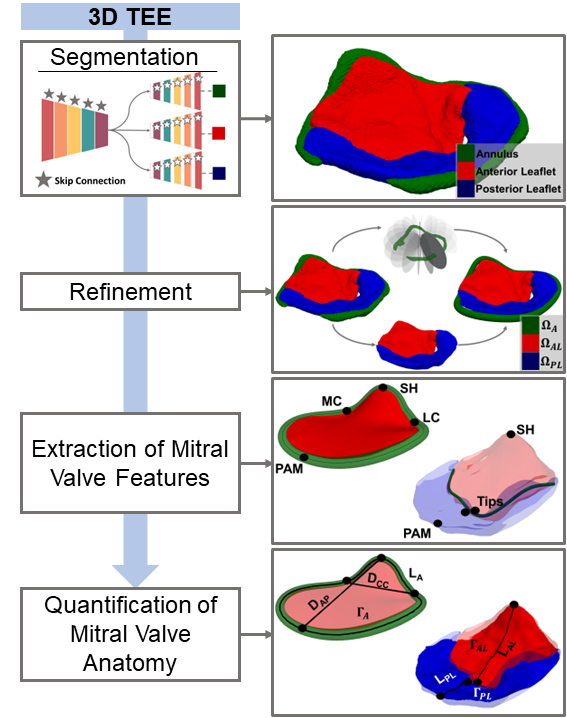}}
 \caption{Schematic representation of the implemented automatic pipeline. SH = saddle horn, PAM = posterior annulus mid-point, MC = medial commissure, LC = lateral commissure, $D_{CC}$ = inter-commissural diameter, $D_{AP}$ = antero-posterior diameter, $L_{A}$ = annulus length, $\bm \Gamma_{A}$= annulus surface, Tips = leaflets tips $L_{PL}$ = posterior leaflet length, $L_{AL}$ = anterior leaflet length, $\bm \Gamma_{AL}$ = anterior leaflet surface, $\bm \Gamma_{PL}$ = posterior leaflet surface.}
 \label{fig:pipeline}
\end{figure}

A fully automatic pipeline (Figure \ref{fig:pipeline}) was implemented, embedding the trained 3D CNN, for \gls{mv} segmentation and morphological characterization. The 3D CNN analyzes \gls{3dtee} images at end-systole and segments the \gls{mv} by separately recognizing the mitral annulus, the anterior leaflet and the posterior leaflet. Given the multi-class segmentation mask provided by 3D CNN (Figure \ref{fig:pipeline}, Segmentation), the implemented pipeline allows to (i) smooth and correct the predicted segmentation (Figure \ref{fig:pipeline}, Refinement) (ii) build a model of the coaptation line, (iii) detect \gls{mv} anatomical landmarks (Figure \ref{fig:pipeline}, extract \gls{mv} features), and (iv) quantify \gls{mv} anatomy (Figure \ref{fig:pipeline}, Quantification of \gls{mv} Anatomy).

\subsubsection*{Smoothing and Correction}

\begin{figure}[ht]
 \centering
 \includegraphics[width=\columnwidth]{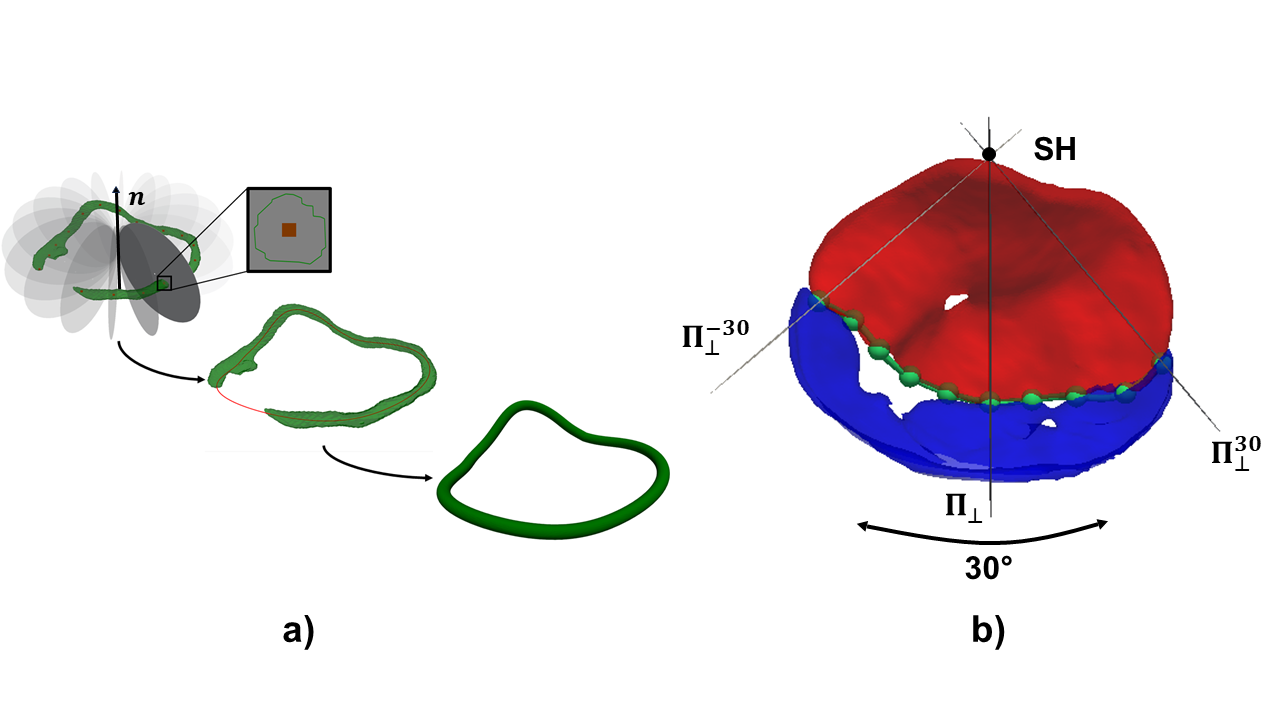}
 \caption{Schematic representation of the algorithms for the annulus reconstruction and coaptation line identification. (a) From the top left to the bottom right, the three steps involved in the annulus reconstruction: 1) identification of intersecting points between $\bm \Omega_{A}$ (green) and the multiple rotating planes (gray); 2) interpolation of the intersecting points for the annulus skeleton (yellow) reconstruction; 3) radial expansion of the annulus skeleton. (b) The reference plane $\bm \Pi_{\perp}$ and two representative examples of the multiple rotating planes $\bm \Pi_{\perp}^{30}$ and $\bm \Pi_{\perp}^{-30}$ (gray); the intersecting points (green) between $\bm \Pi_{\perp}^{i}$ and  $\bm \Omega\textsubscript{AL}$.}
 \label{fig:correction-alg}
\end{figure}

A marching cubes algorithm \cite{Lorensen_1987} is applied to the multi-label segmentation mask to extract three triangulated surface meshes of the mitral annulus $\Omega\textsubscript{A}$, and of the anterior $\Omega\textsubscript{AL}$ and posterior leaflet $\Omega\textsubscript{PL}$, respectively (Figure \ref{fig:pipeline}, Refinement). The surface meshes are then smoothed using a windowed sinc filter \cite{Pinter_2019}. 
To correct for possible discontinuities in annulus segmentation, the skeleton of $\Omega\textsubscript{A}$ is reconstructed and processed to guarantee a closed profile. Briefly, the valve orifice center \textbf{x\textsubscript{c}} is calculated by taking the average of the coordinates \textbf{x}\textsubscript{i} = (\textit{x}\textsubscript{i}, \textit{y}\textsubscript{i}, \textit{z}\textsubscript{i}) of the $n$ points in $\Omega\textsubscript{A}$. Singular-value decomposition (SVD) on the set of directions $\textit{D}:\{\textbf{d}\textsubscript{i}=\textbf{x}\textsubscript{i}-\textbf{x\textsubscript{c}},\quad\textbf{x}\textsubscript{i} \in \Omega\textsubscript{MA} \}$ allows defining the principal directions of the annulus geometry. The radial \textbf{r} and normal \textbf{n} directions are identified through SVD as the direction of highest and lowest variance, respectively. \textbf{x\textsubscript{c}} and \textbf{n} uniquely identify a bounded plane $\Pi$, which is used to define the \gls{mv} orifice plane in the subsequent steps of the automated pipeline. Successively, multiple-rotating planes are defined around \textbf{n} evenly spaced out by an angle $\theta_{offset}=15^{\circ}$. The intersection between these planes and $\Omega\textsubscript{A}$ identifies a set of 3D points \textit{S} outlining the mitral annulus shape (Alg. \ref{alg:Alg1}). Finally, these points are interpolated by a cubic spline that is expanded radially by 1 mm to obtain the refined version of $\Omega\textsubscript{A}$, consistently with the contour radius used for the GT label (Figure \ref{fig:correction-alg}a).

\begin{algorithm}
	\caption{Definition of skeleton of the mitral annulus from segmentation mask } 
	\begin{algorithmic}[H]
	    \STATE Given $\Omega_{A}$ and \textit{D}
	    \STATE $\theta_{offset}=15^{\circ}$, $\theta=0^{\circ}$
	    \STATE \textbf{n} and \textbf{r} $\leftarrow$ SVD(\textit{D})
		\WHILE {$\theta < 2\pi$}
			\STATE \textbf{r}\textsubscript{$\theta$} $\leftarrow$ Rot\textsubscript{$\theta_{offset}$}(\textbf{r})
			\STATE $\Pi$ $\leftarrow$ $\textbf{n}\times\textbf{r}\textsubscript{$\theta$}$
			\STATE Find set of intersecting points \textit{S} between $\Pi$ and $\Omega_{A}$
			\STATE $\theta\leftarrow\theta_{offset} + \theta$
		\ENDWHILE
	\end{algorithmic}
	\label{alg:Alg1}
\end{algorithm}

\subsubsection*{Coaptation Line Identification}

To automatically identify the coaptation line in a repeatable and robust way, the following steps are performed. First, the SH is identified as the highest point of $\Omega\textsubscript{A}$ with respect to $\Pi$. Then, a plane denoted $\Pi_{\perp}$ is defined as the one passing through SH and through \textbf{x\textsubscript{c}}, and perpendicular to $\Pi$. Subsequently, $\Pi_{\perp}$ is iteratively rotated by $5^{\circ}$ around an axis passing through the SH and orthogonal to $\Pi$, towards the medial and lateral portion of the valve  until a total angle of $30^{\circ}$ is spanned for each portion of the valve. This rotation identifies several rotating planes around \textbf{n} denoted as $\Pi_{\perp}^{i}$, where $i$ ranges from -30 to 30, as exemplified in Figure \ref{fig:correction-alg}b.  The candidate points belonging to the coaptation zone are the points with the maximum distance to SH among the set of points identified by the intersection between $\Pi_{\perp}^{i}$ and $\Omega\textsubscript{AL}$ (Figure \ref{fig:correction-alg}b, in green). Finally, a least-square fit of the candidate points with 3\textsuperscript{th} degree polynomial is used to obtain a smooth and continuous line.

\subsubsection*{Detection of Mitral Valve Anatomical Landmarks}

\begin{figure}[ht]
 \centering
 \centerline{\includegraphics[width=\columnwidth]{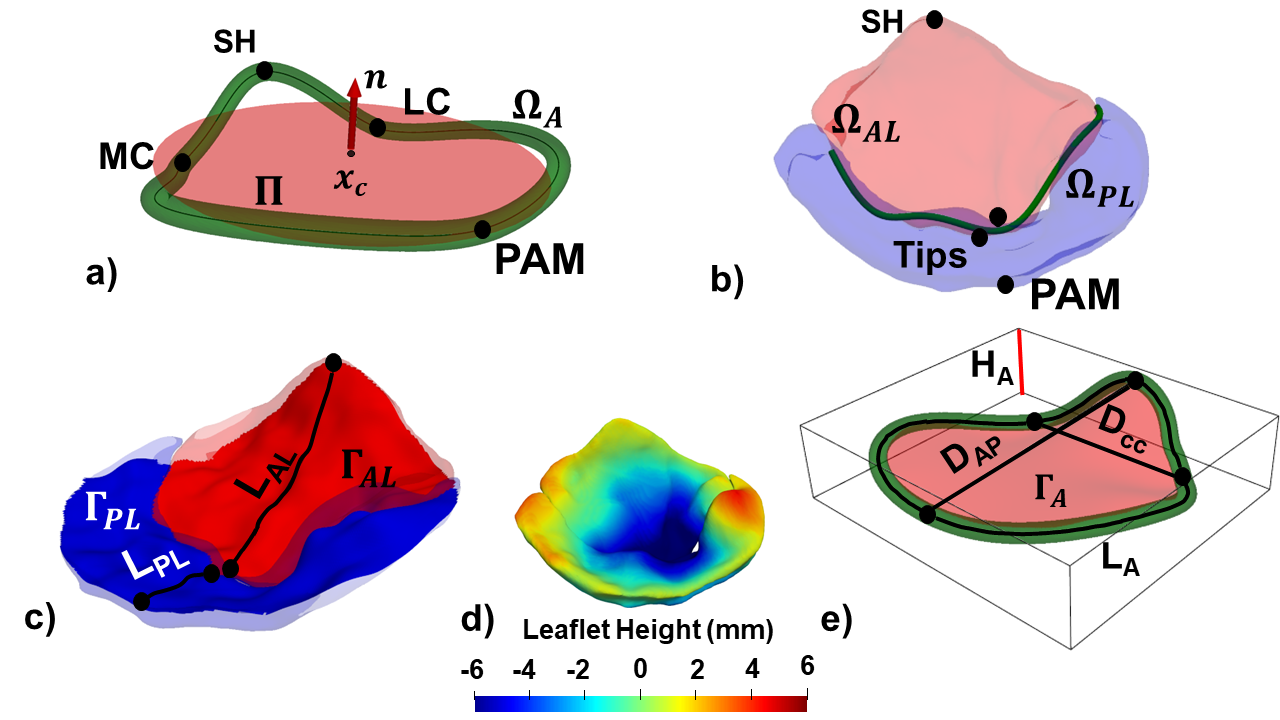}}
 \caption{Illustrative example of reconstructed \gls{mv} model with anatomical features and measurements extracted by the automatic pipeline. (a) Reconstructed and refined mitral annulus ($\bm \Omega_{A}$) together with the annular anatomical landmarks (SH = saddle horn, PAM = posterior annulus mid-point, MC = medial commissure, LC = lateral commissure), the best fitting plane ($\bm \Pi$) and the unitary vector (\textbf{n}) normal to $\bm \Pi$. (b) Reconstructed mitral leaflets ($\bm \Omega_{AL}$=anterior leaflet surface, in red; $\bm \Omega_{PL}$=posterior leaflet surface, in blue), main leaflet landmarks (Tips = leaflets tips), and  the reconstructed model of the coaptation line (green). (c) Leaflet 3D middle surfaces ($\bm \Gamma_{AL}$, $\bm \Gamma_{PL}$) and leaflet 3D measurements ($L_{PL}$ = posterior leaflet length, $L_{AL}$ = anterior leaflet length). (d) Color-coded 3D representation of the reconstructed \gls{mv} representing leaflet height. (e) 3D surface interpolating the non-planar annular profile ($\bm \Gamma_{A}$) and defining the annular area, and annulus 3D measurements ($D_{CC}$ = inter-commissural diameter, $D_{AP}$ = antero-posterior diameter, $L_{A}$ = annulus length, $H_{A}$ = annulus height).}
 \label{fig:output-pipeline}
\end{figure}

The pipeline automatically identifies the \gls{mv} anatomical landmarks commonly used to describe annulus and leaflet anatomy. This is done using the reconstructed surface meshes $\Omega\textsubscript{A}$, $\Omega\textsubscript{AL}$, and $\Omega\textsubscript{PL}$.\\
On the annulus, SH is identified as described in the previous subparagraph. The two commissures (LC, MC) are identified as the closest points of the skeleton of $\Omega\textsubscript{A}$ with respect to the two extremities of the reconstructed coaptation line, while the posterior annular midpoint (PAM) is defined as the point opposite to SH in the posterior portion of $\Gamma\textsubscript{A}$ skeleton (Figure \ref{fig:output-pipeline}a).\\
A plane ($\Pi\textsubscript{SH-PAM}$) is defined by intersecting SH and PAM and being perpendicular to $\Pi$. This plane intersects with the surfaces $\Omega\textsubscript{AL}$ and $\Omega\textsubscript{PL}$, resulting in two sets of points on the leaflet surface. Among these points, the leaflet tips, depicted in Figure \ref{fig:output-pipeline}b, are identified as the two points farthest from SH and PAM.

\subsubsection*{Quantification of Mitral Valve Anatomy}

The final step of the implemented pipeline consists in the automatic computation of \gls{mv} morphological metrics based on the identified anatomical landmarks. For the annulus, the anterior-posterior diameter ($D_{AP}$) and the inter-commissural diameter ($D_{CC}$) are computed as the Euclidean distance from SH to PAM and from MC to LC, respectively. Annular length (L\textsubscript{A}) is defined as the length of the $\Omega\textsubscript{MA}$ skeleton. Finally, the height of the annulus (H\textsubscript{A}) is defined as the height of the smallest hexahedral bounding box encompassing $\Omega\textsubscript{MA}$. Also, a 3D surface ($\Gamma\textsubscript{A}$) representing the non-planar \gls{mv} orifice is reconstructed using a thin plate radial basis function interpolation \cite{Wright_2003} of $\Omega\textsubscript{MA}$. Annular area is defined by as the surface area of $\Gamma\textsubscript{A}$ (Figure \ref{fig:output-pipeline}e). \\
For the leaflets, the length is defined as the length of the intersection between each leaflet and $\Pi\textsubscript{SH-PAM}$, spanning from the SH to the anterior tip and from PAM to the posterior tip, respectively. Two 3D middle surfaces $\Gamma\textsubscript{AL}$ and $\Gamma\textsubscript{PL}$ are defined using a 5\textsuperscript{th} order spline radial basis function (RBF) interpolation \cite{Wright_2003} of $\Omega\textsubscript{AL}$ and $\Omega\textsubscript{PL}$, respectively. Leaflet area is defined as the surface area of $\Gamma\textsubscript{AL}$ and $\Gamma\textsubscript{PL}$ (Figure \ref{fig:output-pipeline}c). The local height of the leaflets is computed as the signed distance of $\Gamma\textsubscript{AL}$ and $\Gamma\textsubscript{PL}$, respectively, from the 3D orifice surface $\Gamma\textsubscript{A}$  (Figure \ref{fig:output-pipeline}d). Leaflet flail or prolapse would result in a greater positive distance from $\Gamma\textsubscript{A}$.

\subsection{Performance Evaluation}
\label{subsec:performance-evaluation}
For each of the ten validation sets (one for each cross-validation fold), Dice score \cite{Zijdenbos_1994}, mean surface distance (MSD, in mm) and 95\% Hausdorff Distance (95\% HD, in mm) \cite{Huttenlocher_1993} were computed for each label, i.e., anterior leaflet, posterior leaflet, and annulus, and for the combined segmentation mask, i.e., the valve as a whole.  Given the points \textbf{p} and \textbf{p'} belonging to the GT surface (S) and to the surface of the predicted segmentation (S'), respectively, MSD was defined as: 

\begin{equation} 
    MSD = \frac{1}{n_{S} + n_{S'}}(\sum_{}^{n_{S}}D(\mathbf{p}, S') + \sum_{}^{n_{S'}}D(\mathbf{p'}, S))  
    \label{Eq:MSD} 
\end{equation} 
         
where $D(\mathbf{p}, S')=\min_{\mathbf{p'} \in S'}\Vert \mathbf{p} - \mathbf{p'} \Vert$ is the Euclidean norm. In the results section, the average performance metrics of the ten training-validation splits are shown. \\
The ten trained models were ensembled and evaluated on an independent test set comprising 5 3DTEE volumes. Dice score, MSD and 95\% HD were calculated to evaluate the performance of the ensemble models. The MSD and 95\% HD were recomputed after applying the correction algorithm (Figure 3 in Section II-E). The refinement module within the proposed pipeline operates exclusively on the surface extracted from the segmentation. Consequently, post-refinement computation of the Dice score, reliant on image analysis, becomes impractical. The ensembled model was then integrated within the proposed pipeline to extract the relevant anatomical features of the \gls{mv} and quantify its anatomy on the test set. To validate the proposed automatic morphological analysis, results were compared against measurements obtained using TomTec Image Arena, a widely recognized commercial software known for its semi-automated tool for \gls{mv} modelling (4D \gls{mv} Assessment, version 4.6). This software is extensively utilized in clinical settings for evaluating the morphology of diseased \gls{mv}s. TomTec's tool enables the creation of a \gls{mv} leaflets triangulated surface mesh from \gls{3dtee} data, following initialization steps provided by the user. The triangulated surface mesh provided by TomTec is comprehensive of \gls{mv} 3D shape, but it lacks information about leaflets’ thickness, presenting the surface as a topological 2D object in a 3D space. Morphological assessments of the \gls{mv} are then derived from this surface. Specifically, as described in \cite{Morbach_2018}, this tool requires the user to perform a specific sequence of actions to create a 3D surface mesh of the \gls{mv}. Briefly, the user must trace two points representing the annulus on each of two automatically selected 2D views, i.e., the four-chamber and the two-chamber view, which are on mutually orthogonal planes. As a second step, the aortic valve must be identified by manually orienting a long axis view; finally, once the aortic valve is identified on the long axis view, the user must trace the SH and the coaptation point in the same long axis view. Based on this initialization, the software automatically generates the model of \gls{mv} leaflets surface and of \gls{mv} annulus, which can be manually refined by the user. The software then automatically identifies the anatomical landmarks commonly used to describe the anatomy of \gls{mv} annulus and leaflets, and based on these landmarks calculates the corresponding measurements (\cite{Morbach_2018, Aruta_2018}: diameters, perimeter, 2D and 3D areas and height of the annulus; length and surface area of the leaflets. Although it is widely used in clinical settings, its reliance on manual inputs might introduce some biases in the morphological measurements. To address potential biases, morphological measurements extracted by the proposed automatic morphological analysis from the GT segmentations were also considered. The analysis was executed by comparing the morphology measurements computed by our proposed method, based on the surfaces of \gls{mv} leaflets and annulus extracted from the CNN-based segmentation after refinement, against those provided by TomTec. Notably, no post-processing was applied to TomTec-derived measurements before this comparison.\\ 
To test the benefit brought by the proposed multi-decoder U-Net architecture for \gls{mv} segmentation from \gls{3dtee} images, a baseline Residual U-Net with an equivalent encoder and a single decoder \cite{Carnahan_2021} was trained adopting the same training strategy (10-fold cross validation and model ensembling). The performance of the ensembled models based on the two architectures were compared on the test set, computing the Dice score, MSD and 95\% HD. Upon verifying the normal distribution of data through a Shapiro-Wilk normality test, a two-sample t-test was run to test the statistical significance of the difference in Dice Score, MSD, and 95\% HD between the proposed model and the basic Residual U-Net. Differences were deemed statistically significant for p-value<0.05.  Because the available test set was relatively small, the analysis was conducted on the average of the results of the 10 validation folds to have a statistically significant number of data, since a small number of data limits the interpretation of this statistical test.\\
All performance metrics are expressed as mean value $\pm$ standard deviation.

\section{Results}
\subsection{Inter-operator variability}
\label{subsec:inter-operator variability}
For the 10 randomly selected \gls{3dtee} images, the MV segmentations obtained by the three independent operators was compared pairwise, in terms of Dice Score, MSD and 95\% HD considering the complete segmentation mask. The results of this comparison are reported in Table \ref{tab:tabel-1}. Overall, good agreement was observed between the MV segmentations by any two operators, with mean Dice score values ranging from 0.75 to 0.83. On average, the proposed protocol took 11.14 $\pm$ 3.04 min to segment MV in a single \gls{3dtee} volume.

\begin{table}[ht]
\begin{tabular}{lccc}
\hline
                & Op1 vs. Op2      & Op1 vs. Op3     & Op2 vs. Op3     \\ \hline
Dice            & 0.83 $\pm$ 0.02  & 0.79 $\pm$ 0.04 & 0.75 $\pm$ 0.04 \\
MSD {[}mm{]}    & 0.33 $\pm$ 0.06  & 0.39 $\pm$ 0.07 & 0.48 $\pm$ 0.10 \\
95\%HD {[}mm{]} & 2.72 $\pm$ 0.44 & 4.03 $\pm$ 1.20 & 4.57 $\pm$ 1.25 \\ \hline
\end{tabular}
\caption{Inter-operator variability analysis results. Dice score, MSD and 95\% HD are computed for each pairwise comparison. Mean values $\pm$ standard deviations are reported.}
\label{tab:tabel-1}
\end{table}

\subsection{Test Set Results}
\label{subsec:test-set-results}
\begin{figure*}[ht]
 \centering
 \centerline{\includegraphics[width=\textwidth]{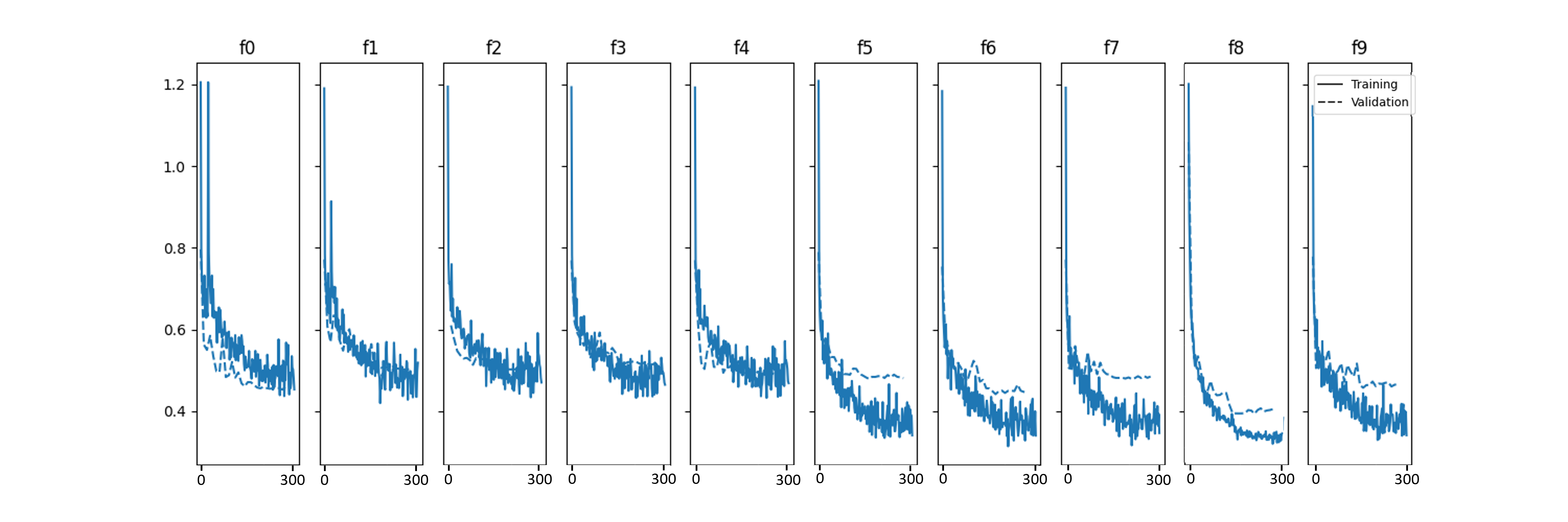}}
 \caption{Training and validation loss curves for each validation subset (f0-f9) over the total amount of epochs. Continuous line represents the training loss while the dashed line represents the validation loss.}
 \label{fig:training and validation curves}
\end{figure*}

\begin{figure}[ht]
    \centering
    \includegraphics[width=\columnwidth]{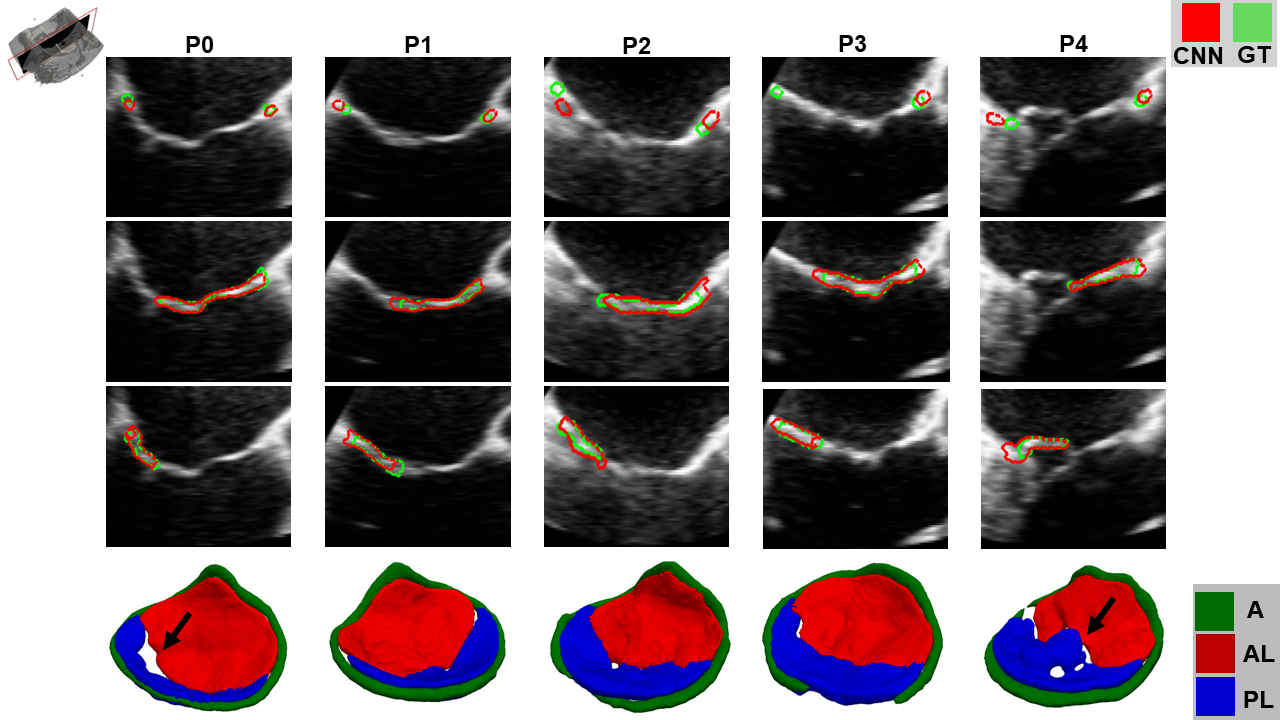}
    \caption{Long-axis views of 3D TEE images and segmentation masks for each volume in test set. In the first 3 rows the GT label is shown in green, and predicted label (CNN) is shown in red for annulus, anterior leaflet and posterior leaflet, respectively. In the fourth row, a 3D representation of predicted raw segmentation mask is shown, using a color code to distinguish annulus (A), anterior leaflet (AL) and posterior leaflet (PL). Black arrows on 3D representations highlight leaflet defects (P0) or prolapse (P4) correctly identified in the predicted segmentation masks.} 
    \label{fig:test-set}
\end{figure}

\begin{figure}[ht]
    \centering
    \includegraphics[width=\columnwidth]{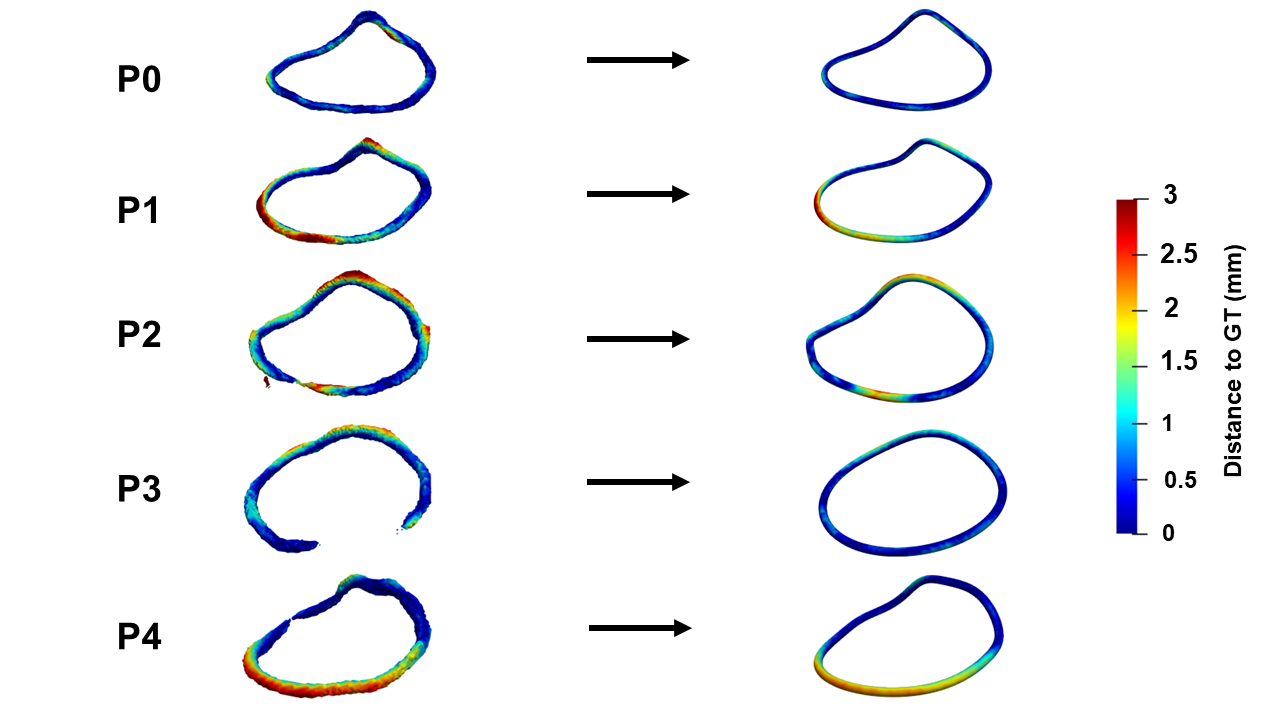}
    \caption{Heatmaps of the distance of the annulus from the GT reference for the 5 volumes of the testing set. Data are reported for the annulus before (left column) and after (right column) the smoothing and correction steps of the proposed pipeline.} 
    \label{fig:surface-distance}
\end{figure}

Figure \ref{fig:training and validation curves} illustrates the training and validation curves for each of the 10 validation subsets. Among the first five subsets (f0-f4), the models consistently exhibited behaviours that do not suggest overfitting. However, in the remaining subsets (f5-f9), the validation curves marginally surpassed the training curves. This trend implies a potential lack of generalization in these folds, likely stemming from the constraints of our limited dataset. This observed behaviour justifies the adoption of an ensemble strategy to evaluate the test set. By amalgamating diverse perspectives from individual models, the potential lack of generalization is mitigated. The average evaluation metrics across these subsets, encompassing annulus, anterior leaflet, posterior leaflet, and the complete segmentation mask, are detailed in Table \ref{tab:tabel-2}.\\
The evaluation metrics achieved by the ensembled model on the test set are reported in Table \ref{tab:tabel-3}. Overall, looking at the average values across the test set (Table \ref{tab:tabel-3}, last row) a significant improvement in the performance metrics was obtained by ensembling the 10 models, especially in MSD and 95\% HD. The ensembled model was capable of accurately extracting the \gls{mv} substructures' anatomy (as shown in Figure \ref{fig:test-set}). For all 5 volumes used for testing, the annulus, and leaflets were successfully segmented in accordance with the GT segmentation masks. Notably, the predicted segmentation masks identified a coaptation defect (P0 in Figure \ref{fig:test-set}) and a \gls{mvp} (P4 in Figure \ref{fig:test-set}) in two test volumes, respectively. In three test cases (P2, P3 and P4 in Figure \ref{fig:test-set}) the predicted mitral annulus presented discontinuities. Overall, the predicted segmentation masks were in agreement with the corresponding evaluation metrics. Notably, the annulus label was the most challenging substructure to identify with accuracy, as reflected by the lower performance metrics with respect to the anterior and posterior labels. Specifically, the predicted annulus achieved an averaged Dice score of 0.40 $\pm$ 0.10, which was lower than the average Dice score of 0.81 $\pm$ 0.33 for the predicted anterior leaflet and 0.68 $\pm$ 0.07 for the predicted posterior leaflet. \\
The recalculated MSD and 95\% HD errors, post-application of the refinement algorithm (Figure \ref{fig:correction-alg} in Section \ref{subsec:pipeline-implementation}) for all five test volumes, are detailed in Table \ref{tab:tabel-4}. Upon comparing Table 3 with Table 4, there were no significant improvements observed for the labels corresponding to the leaflets and the complete segmentation mask in terms of distance metrics, between pre- and post-refinement. However, after refinement, the annulus label exhibited an average MSD of 0.85 ± 0.35 mm and a 95\% HD of 3.68 ± 1.13 mm (as shown in Table \ref{tab:tabel-4}). These values signify an enhancement of 0.10 mm in the MSD and 2.97 mm in the 95\% HD when compared to the original measurements listed in Table 3. In all five test volumes, the refinement of the inferred annulus segmentations yielded coherent annular profiles with respect to the GT annotations, also in those cases where the raw annulus segmentation was discontinuous or incomplete, and reduced the surface error (Figure \ref{fig:surface-distance}).

\begin{table}[ht]
    \centering
    \resizebox{\columnwidth}{!}{%
    \begin{tabular}{lccc}
    \hline
                  & \multicolumn{3}{c}{Average Performance Metrics of 10-Fold Cross Validation} \\
                  & Dice Score              & MSD {[}mm{]}           & 95\% HD {[}mm{]}           \\ \hline
    Annulus       & 0.41 $\pm$ 0.04   & 2.57 $\pm$ 1.64  & 16.84 $\pm$ 6.47  \\
    Anterior      & 0.77 $\pm$ 0.04   & 1.06 $\pm$ 0.89  & 7.32 $\pm$ 3.58   \\
    Posterior     & 0.68 $\pm$ 0.05   & 1.43 $\pm$ 0.86  & 9.92 $\pm$ 2.34   \\
    Complete Mask & 0.81 $\pm$ 0.02   & 0.84 $\pm$ 0.66  & 4.83 $\pm$ 1.56    \\ \hline
    \end{tabular}}
    \caption{Average Dice score, MSD and 95\% HD across the training-validation splits of the 10-fold cross validation. Performance metrics are reported for mitral annulus (Annulus), anterior leaflet (Anterior), posterior leaflet (Posterior) and for complete segmentation mask (Complete Mask). Mean values $\pm$ standard deviations are reported.}
    \label{tab:tabel-2}
    \end{table}

\begin{table*}[ht]
    \centering
    \resizebox{\textwidth}{!}{%
    \begin{tabular}{lcccccccccccc}
    \hline
            & \multicolumn{3}{c}{Annulus}                                    & \multicolumn{3}{c}{Anterior}                                   & \multicolumn{3}{c}{Posterior}                                  & \multicolumn{3}{c}{Complete Mask}                             \\
            & Dice Score & \multicolumn{1}{l}{MSD {[}mm{]}} & 95\% HD {[}mm{]} & Dice Score & MSD {[}mm{]} & \multicolumn{1}{l}{95\% HD {[}mm{]}} & Dice Score & MSD {[}mm{]} & \multicolumn{1}{l}{95\% HD {[}mm{]}} & Dice Score & \multicolumn{1}{l}{MSD {[}mm{]}} & 95\% HD{[}mm{]} \\ \hline
    P0      & 0.56       & 0.54                             & 3.20           & 0.86       & 0.31         & 2.23                               & 0.75       & 0.44         & 4.12                               & 0.88       & 0.32                             & 2.00          \\
    P1      & 0.39       & 0.93                             & 4.15           & 0.80       & 0.51         & 6.40                               & 0.69       & 0.82         & 7.81                               & 0.83       & 0.41                             & 3.74          \\
    P2      & 0.35       & 1.03                             & 12.50           & 0.76       & 0.46         & 5.09                               & 0.69       & 0.54         & 7.87                               & 0.81       & 0.45                            & 3.46          \\
    P3      & 0.47       & 0.86                             & 8.40          & 0.79       & 0.59         & 8.54                               & 0.74       & 0.73         & 7.54                               & 0.88       & 0.33                             & 2.23          \\
    P4      & 0.24       & 1.38                             & 5.02           & 0.81       & 0.43         & 4.47                               & 0.56       & 1.05         & 7.81                               & 0.72       & 0.67                             & 6.4          \\
    Average & 0.40 $\pm$ 0.10  & 0.95 $\pm$ 0.27                        & 6.65 $\pm$ 3.81      & 0.81 $\pm$ 0.33  & 0.46 $\pm$ 0.09    & 5.35 $\pm$ 2.09                          & 0.68 $\pm$ 0.07  & 0.72 $\pm$ 0.21    & 7.03 $\pm$ 1.45                          & 0.82 $\pm$ 0.06  & 0.43 $\pm$ 0.14                        & 3.57 $\pm$ 1.56     \\ \hline
    \end{tabular}}
    \caption{Dice score, MSD and 95\% HD for each volume of test set. Performance metrics are reported for mitral annulus (Annulus), anterior leaflet (Anterior), posterior leaflet (Posterior) and for complete segmentation mask (Complete Mask). Average value across the test set and  standard deviation are given in the last row.}
    \label{tab:tabel-3}
\end{table*}

\begin{table*}[ht]
    \centering
    \resizebox{\textwidth}{!}{%
    \begin{tabular}{lcccccccc}
    \hline
            & \multicolumn{2}{c}{Annulus}                                    & \multicolumn{2}{c}{Anterior}                                   & \multicolumn{2}{c}{Posterior}                                  & \multicolumn{2}{c}{Complete Mask}                             \\
            &  \multicolumn{1}{l}{MSD {[}mm{]}} & 95\% HD {[}mm{]} & MSD {[}mm{]} & \multicolumn{1}{l}{95\% HD {[}mm{]}} & MSD {[}mm{]} & \multicolumn{1}{l}{95\% HD {[}mm{]}} & \multicolumn{1}{l}{MSD {[}mm{]}} & 95\% HD{[}mm{]} \\ \hline
    P0        & 0.51                             & 2.47              & 0.32         & 3.73                                  & 0.45         & 3.67                                     & 0.34                             & 2.48          \\
    P1            & 0.90                             & 4.10                 & 0.51         & 6.57                                   & 0.83         & 8.65                                    & 0.44                             & 3.33          \\
    P2            & 0.94                             & 4.32                 & 0.46         & 4.08                                    & 0.55         & 4.64                                    & 0.45                            & 3.02          \\
    P3            & 0.54                             & 2.50                & 0.59         & 7.92                                     & 0.73         & 6.98                                     & 0.32                             & 1.79          \\
    P4            & 1.38                             & 5.00                 & 0.44         & 5.84                                     & 1.05         & 4.84                                     & 0.68                             & 4.60          \\
    Average   & 0.85 $\pm$ 0.35                        & 3.68 $\pm$ 1.13       & 0.46 $\pm$ 0.10    & 5.63 $\pm$ 1.74                           & 0.72 $\pm$ 0.23    & 5.76 $\pm$ 2.02                       & 0.45 $\pm$ 0.14                        & 3.04 $\pm$ 1.04     \\ \hline
    \end{tabular}}
    \caption{MSD and 95\% HD for each volume of test set after applying the refinement module. Performance metrics are reported for mitral annulus (Annulus), anterior leaflet (Anterior), posterior leaflet (Posterior) and for complete segmentation mask (Complete Mask). Average value across the test set and  standard deviation are given in the last row.}
    \label{tab:tabel-4}
\end{table*}

\subsection{Comparison vs. Semi-Automated Measurements}
\label{automatic-analysis}
Table \ref{tab:table-5} provides a comparative analysis of average morphological measurements across the test set obtained by the proposed method, alongside measurements derived from GT segmentations and those generated by TomTec's semi-automated 4D \gls{mv} Assessment tool. The table illustrates the average differences between the GT and the proposed method, as well as between the GT and TomTec. Overall, good agreement was observed for geometric measurements between the proposed method and the GT, consistent with the findings in Tables \ref{tab:tabel-3} and \ref{tab:tabel-4}. However, some small deviations were noted: the proposed method slightly overestimated the inter-commissural diameter, anteroposterior diameter, anterior and posterior leaflet length, with the maximum shift being 1.45 mm. The largest shift of 6.58 mm was observed for the annulus length. Regarding area estimations, overall consistency was observed, with the largest bias of 105.64 mm\textsuperscript{2} for the anterior leaflet surface area. Comparatively, differences with TomTec's measurements were marginally higher in absolute terms, but no specific trend was evident. TomTec tended to underestimate the annulus area, exhibiting an average bias of 192.76 mm\textsuperscript{2} compared to the GT. In general, the commercial software required considerable user interaction, taking about 3 minutes for an experienced user to process a single \gls{3dtee} volume, on average.

\begin{table*}[ht]
    \resizebox{\textwidth}{!}{%
    \begin{tabular}{lccccccc}
        \hline
                                            & Proposed & Ground Truth & TomTec  & Bias (Proposed vs. GT) & 95\% limits of agreement (Proposed vs. GT) & Bias (TomTec vs. GT) & 95\% limits of agreement (TomTec vs. GT) \\ \hline
        Annulus                             &          &              &         &                                  &                                                      &                                &                                                    \\
        Inter-commisural diameter, mm       & 48.46    & 47.01        & 45.14   & -1.45                            & {[}-3.80, 0.91{]}                                    & 1.87                           & {[}-0.87, 4.62{]}                                  \\
        Anteroposterior diameter, mm        & 39.36    & 38.17        & 40.12   & -1.19                            & {[}-4.80, 2.41{]}                                    & -1.95                          & {[}-6.67, 2.77{]}                                  \\
        Height, mm                          & 13.55    & 13.71        & 10.38   & 0.16                             & {[}-2.12, 2.45{]}                                    & 3.33                           & {[}-0.55, 7.22{]}                                  \\
        Length, mm                          & 145.84   & 139.26       & 142.62  & -6.58                            & {[}-16.21, 3.05{]}                                   & -3.35                          & {[}-12.47, 5.77{]}                                 \\
        Area, mm2                           & 1498.80  & 1556.30      & 1748.76 & -57.5                            & {[}-384.92, 252.93{]}                                & -192.76                        & {[}-300.17, 79.93{]}                               \\
        Leaflet                             &          &              &         &                                  &                                                      &                                &                                                    \\
        Anterior leaflet length, mm         & 28.08    & 25.98        & 33.55   & -2.10                            & {[}-7.82, 3.62{]}                                    & -7.57                          & {[}-16.03, 1.00{]}                                 \\
        Posterior leaflet length, mm        & 14.47    & 14.33        & 15.02   & -0.14                            & {[}-5.09. 4.81{]}                                    & -0.69                          & {[}-9.75, 8.38{]}                                  \\
        Anterior leaflet surface area, mm2  & 1109.45  & 1003.81      & 1129.20 & -105.64                          & {[}-350.30, 139.02{]}                                & -125.38                        & {[}-247.77, 498.54{]}                              \\
        Posterior leaflet surface area, mm2 & 720.90   & 653.43       & 638.80  & -67.47                           & {[}-279.42, 144.49{]}                                & 14.43                          & {[}-250.82, 311.55{]}                              \\ \hline              
    \end{tabular}}
    \caption{3D \gls{mv} morphology analysis. Comparison between the average measurements across the 5 3DTEE volumes in the test set of annulus and leaflet geometry obtained with the proposed automated pipeline, the GT segmentation masks and TomTec. Mean differences (Bias) and 95\% limits of agreement are reported between the proposed method and GT and TomTec and GT.}
    \label{tab:table-5}
\end{table*}

\subsection{Comparison with Baseline Residual U-Net}
\label{subsec:Comparison-residualunet}
A baseline Residual U-Net architecture was trained by adopting the same training strategy used for the multi-decoder U-Net architecture (10-fold cross validation and model ensembling). Table \ref{tab:tabel-5} compares the average performance metrics obtained by the baseline Residual U-Net and the multi-decoder U-Net on test set. The results indicate that there was no significant difference in the average Dice Score between the two architectures. On the other hand, the proposed multi-decoder U-Net architecture achieved notable improvements in terms of average MSD compared to the baseline Residual U-Net. It resulted in an average MSD reduction of 0.35 mm, 0.39 mm, and 0.30 mm for the annulus, anterior leaflet, and posterior leaflet, respectively, when compared to the average MSD achieved by the baseline Residual U-Net (Table \ref{tab:tabel-5}). Additionally, the 95\% HD also showed significant improvement when using the multi-decoder U-Net, particularly for the annulus label, with a reduction from 19.26 mm to 6.65 mm.\\
The p-values for Dice score, MSD and 95\% HD for the combined segmentation masks were 0.01, 0.03 and 0.2, respectively. Although significant differences were observed only for Dice score and MSD, the multi-decoder U-Net achieved a better value and lower for all the performance metrics with respect to a baseline Residual U-Net. Overall, although this comparison was made on a small dataset, the results obtained suggest that the quality of multi-label segmentation masks may benefit from the use of a multi-decoder U-Net architecture.
    
\begin{table}[ht]
    \centering
    \resizebox{\columnwidth}{!}{%
    \begin{tabular}{lcccccc}
    \hline
             & \multicolumn{3}{c}{Multi-decoder U-Net} & \multicolumn{3}{c}{Baseline U-Net}                                      \\
                  & Dice Score            & MSD {[}mm{]}          & 95\% HD {[}mm{]}        & Dice Score            & MSD {[}mm{]}          & 95\% HD {[}mm{]}          \\ \hline
    Annulus       & 0.40 $\pm$ 0.10 & 0.95 $\pm$ 0.27 & 6.65 $\pm$ 3.81 & 0.41 $\pm$ 0.09 & 1.30 $\pm$ 0.40 & 19.26 $\pm$ 11.72 \\
    Anterior      & 0.81 $\pm$ 0.33 & 0.46 $\pm$ 0.09 & 5.35 $\pm$ 2.09 & 0.80 $\pm$ 0.02 & 0.85 $\pm$ 0.35 & 5.49 $\pm$ 3.57  \\
    Posterior     & 0.68 $\pm$ 0.07 & 0.72 $\pm$ 0.21 & 7.03 $\pm$ 1.45 & 0.69 $\pm$ 0.07 & 1.02 $\pm$ 0.33 & 5.81 $\pm$ 2.48   \\
    Complete Mask & 0.82 $\pm$ 0.06 & 0.43 $\pm$ 0.14 & 3.57 $\pm$ 1.56 & 0.80 $\pm$ 0.04 & 0.82 $\pm$ 0.26 & 4.01 $\pm$ 1.85   \\ \hline
    \end{tabular}}
    \caption{Dice score, MSD and 95\% HD in mm computed on test set using the Multi-decoder U-Net and the baseline Residual U-Net. Data are expressed as mean value $\pm$ standard deviation.}
    \label{tab:tabel-5}
\end{table}

\subsection{Inference Time}
\label{subsec:inference-time} 
The proposed 3D CNN model takes on average 2.94 $\pm$ 1.89 s to provide a raw multi-label segmentation mask running on a GPU Nvidia RTX A4000. The ensemble of 10 3DCNN models had an impact in the speed performance of the segmentation process, slightly increasing the inference time required as compared to using a single model. The automated pipeline takes on average 11.74  $\pm$ 0.91 s for the morphometric characterization of the \gls{mv} with 1.13 $\pm$ 0.91 s dedicated to refining the segmentation mask. This computation is performed using a CPU-only implementation on an Intel Xenon W-2235. Overall, the proposed pipeline demonstrated to be faster than state-of-the-art commercial software, requiring 14.68 s to process a single \gls{3dtee} volume, on average, whereas $\sim$ 3 minutes were required by an experienced user utilizing the semi-automated tool provided by TomTec Image Arena.

\section{Discussion}
\label{sec:discussion}
A deep learning-based pipeline was presented to automatically segment \gls{mv} annulus and leaflets in end-systolic configuration from \gls{3dtee} images and extract the relevant anatomical landmarks and features to quantitatively assess \gls{mv} anatomy.\\
From a methodological standpoint, the primary novelty of the presented study is the development of a 3D CNN with a multi-decoder residual U-Net architecture that provides multi-structure end-to-end segmentation. The model was trained based on a supervised training strategy where the input data were meticulously labelled by three experienced operators using 3DSlicer, all adhering to the same labelling protocol (subsection \ref{subsec: manual_annotation}). An inter-operator variability analysis was conducted on a randomly selected subset of the available data, which revealed that the GT data exhibited relatively low variability (Dice score 0.85 Op1 vs. Op2, 0.79 Op1 vs. Op3, 0.75 Op2 vs. Op3). This evidence suggests the repeatability and consistency of the proposed protocol for generating GT data. The model was trained using a 10-fold cross-validation approach and evaluated by combining the outputs of the 10 trained models on an independent test set. This ensembling technique was used to improve the overall performance and reduce potential bias. The ensemble of the CNN models achieved an average Dice score of 0.82 (range 0.72-0.88) for the combined segmentation mask, which is comparable to the Dice score (0.81 $\pm$ 0.07) reported in \cite{Carnahan_2021} and slightly lower than the Dice Score (0.877 $\pm$ 0.027) achieved by \cite{Chen_2023}. Nonetheless, the results achieved in this work are overall consistent with the level of agreement observed among manual operators. However, determining the best method remains difficult because of the potential influences of data annotation standards, and of size and quality of datasets. It is worth mentioning that Carnahan et al. \cite{Carnahan_2021} proposed a deep-learning pipeline to segment \gls{mv} leaflets in their open configuration, whereas Chen et al. \cite{Chen_2023} aimed to achieve time-varying \gls{mv} segmentation. Nevertheless, none of the two studies managed to discriminate between the different \gls{mv} substructures. This limitation inherently hinders the identification of the leaflet coaptation line: a key anatomical feature for supporting the identification of coaptation defects and a pivotal landmark during TEER procedures, where the pose of the implantable device that will enforce leaflet coaptation is set by the operator based on the location of the coaptation defect and on the local orientation of the coaptation line to be restored \cite{Nyman_2018}. The proposed ensembling strategy improved the single model performance and increased the robustness of the overall system, at the cost of a slight increase in inference time. As shown in Figure \ref{fig:test-set}, the predicted segmentation masks given by the ensembled model were consistent with the corresponding GT segmentations. The mitral annulus is the label that showed the least overlap with the GT datum, as suggested by the relatively low Dice score obtained for this label. However, it should be pointed out that the annulus is the thinnest structure; thus, small discrepancies in annular profile may reflect in a notable decrease in Dice score. On the other hand, the anterior and posterior leaflets showed a good degree of overlap with the corresponding GT segmentations (average Dice score of 0.81 $\pm$ 0.33 and 0.68 $\pm$ 0.07, respectively), especially along the coaptation region. This result suggests that the presented automated pipeline reliably identifies the coaptation line, coaptation defects (P0 in Figure \ref{fig:test-set}) or \gls{mvp}s (P4 in Figure \ref{fig:test-set}), which are distinctive features of \gls{mv}s considered for percutaneous \gls{mv} repair. The average surface distance to the GT data for the full segmentation mask (0.43 $\pm$ 0.14 mm) was of the order of the spatial resolution of the TEE volumes in the dataset (0.30-0.70 mm/voxel). Furthermore, the reported distance was almost equal to the inter-user variability typical of manual segmentation, previously reported as 0.6 $\pm$ 0.17 mm \cite{Jassar_2011}, and compared favorably with previous semi-automatic (0.59 $\pm$ 0.49 mm \cite{Schneider_2011} and 0.60 $\pm$ 0.20 mm \cite{Pouch_2017}) and fully automated methods (1.54 $\pm$ 1.17 mm \cite{Ionasec_2011}, 0.59 $\pm$ 0.23 mm \cite{Carnahan_2021} 0.925 $\pm$ 0.392 \cite{Chen_2023}). The MSD and 95\% HD were observed to be slightly lower when computed for the complete mask as compared to each label separately. This discrepancy can be attributed to the absence of well-defined intensity-based boundaries for \gls{mv} structures in the \gls{3dtee} images, which makes it challenging to distinguish between them accurately. As a result, any mis-labeled region within each substructure can significantly affect the deterioration of this metric, particularly for smaller structures like the mitral annulus. Furthermore, {the presented} multi-decoder residual U-Net architecture demonstrated superior performance metrics when compared to a baseline residual U-Net for segmenting \gls{mv} from \gls{3dtee} images. Specifically, the proposed method achieved lower values of MSD and 95\% HD for the annulus and leaflets. However no significant difference in terms of Dice score was observed in this comparison (Table \ref{tab:tabel-5}). While the improvement may not be substantial in absolute terms, it is significant when compared to the order of dimension of \gls{mv}. This suggests that both architectures perform similarly in terms of overall segmentation accuracy. Nevertheless, the proposed method excels in distinguishing between various structures by incorporating information from multiple scales and levels of abstraction. It also captures spatial relationships and distances within the data, which is particularly valuable for morphological analysis of \gls{mv} structures in echocardiography. However, it is important to note that this comparison was conducted using a relatively small dataset, and further investigations are needed to validate and confirm these findings.\\
The second key novelty of the study is the introduction of an automatic pipeline that utilizes the CNN-based segmentation mask to characterize the \gls{mv} morphology automatically. The pipeline offers several advantages and achievements. Firstly, the implemented correction algorithm, as illustrated in Figure \ref{fig:surface-distance}, plays a crucial role in ensuring a robust and complete reconstruction of the annulus. This correction algorithm resulted in a significant improvement in performance metrics. Specifically, the MSD and 95\% HD for the predicted annulus were measured at 0.85 $\pm$ 0.35 mm and 3.68 $\pm$ 1.13 mm, respectively. This outperformed the approaches proposed in \cite{Zhang_2020} (2.74 mm) and \cite{Andreassen_2019} (2 mm) for mitral annulus segmentation, showcasing the superior performance of the proposed pipeline. Secondly, the pipeline demonstrated consistent identification of \gls{mv} anatomical landmarks and reconstruction of anatomical features extracted from the CNN-based \gls{mv} segmentation. The comparison between the proposed method and GT-based morphological measurements revealed overall strong agreement. The proposed method slightly overestimated 3D linear measurements by less than 1.45 mm, except for the annulus length, which displayed a 6.58 mm shift. Similarly, area estimations exhibited consistent alignment. These findings align with the segmentation performance metrics discussed earlier, affirming the robustness and accuracy of our proposed method in delivering precise and repeatable \gls{mv} morphology analysis. In contrast, differences with TomTec's measurements were marginally higher in absolute terms. Notably, a bias of 192.76 mm\textsuperscript{2} was identified for the annulus area estimation. This disparity stemmed from TomTec's approach, which delineates the mitral annulus as a closed line representing the outer borders of the mitral leaflets, considered as a single surface. In contrast, the proposed automatic solution represents the \gls{mv} annulus and leaflets with finite thickness, offering a more realistic \gls{mv} depiction. The observed differences in relation to TomTec's measurements can be attributed to the distinct data representation utilized by each method. While TomTec operates on a simplified triangulated surface, the proposed method incorporates a more detailed representation, contributing to the nuanced variations in results. As stated by Chen et al. \cite{Chen_2023}, comparison of morphological measurements extracted from an automatically segmented \gls{mv} model with results from commercial software is further evidence of the accuracy and robustness of the proposed segmentation method. However, unlike \cite{Chen_2023}, the proposed pipeline not only proved able to automate the segmentation of \gls{mv}s from \gls{3dtee} images, but also automates the morphological characterization of regurgitant \gls{mv}s. In this context, the comparison against the commercial software served the purpose of validating the complete tool, including both segmentation and morphological analysis functionalities.\\
Finally, the proposed pipeline is fully automated and requires no manual input at any of its steps. The resulting segmentations and subsequent analyses are therefore be more reproducible than those yielded by manual or semi-automated approaches. The proposed pipeline provided a full and refined \gls{mv} model in 14.68 s on average (range [11.83 s, 19.06 s])  along with a complete morphological characterization; hence, it proved substantially faster than the gold standard method herein considered for comparison, which also required extensive user interaction, making the proposed tool suitable to be used for large population studies where end-to-end automation is a key enabling feature.\\
To the best of our knowledge, none of the previously proposed methods for \gls{mv} segmentation integrated annulus and leaflets quantitative characterization into a single workflow like the pipeline herein presented. As compared to previous CNN-based approaches for segmentation of TEE images that focused only on mitral annulus \cite{Zhang_2020, Andreassen_2019} or leaflets segmentation \cite{Carnahan_2021, Chen_2023}, the 3D CNN presented in this work proved capable to segment and identify all substructures of the \gls{mv}. Moreover, the presented approach addressed the challenging task of segmenting regurgitant \gls{mv}s at end-systole. The characterization of annulus and leaflets anatomy at this time instant allows extracting relevant information to support \gls{mr} diagnosis and \gls{mv} repair planning. 

\subsection{Limitations and Future work}
\label{subsec:limitationsnfuturework}
This study was limited by the difficulty of data collection and labelling. The number of images counted in the dataset is far from the huge dataset used in machine learning studies dealing with natural images. However, the numerosity of the dataset is comparable to that of previous works on automatic echocardiography segmentation. For instance, Carnahan et al. trained a residual U-Net on a set of 48 \gls{3dtee} images \cite{Carnahan_2021}, whereas in the work of Chen et al. the dataset consisted of 44 images \cite{Chen_2023}. Also, unlike Carnahan et al. \cite{Carnahan_2021}, where no data augmentation routine was applied during training, in this study multiple random transformations were applied. This inevitably contributed to increasing the variability of the dataset adopted in this study.\\
The protocol adopted to create the GT data involved several manual actions; it was hence time consuming and potentially operator dependent. The second potential hurdle was assessed through an inter-operator variability analysis. The results revealed that the segmentations produced by the different raters were not entirely congruent, underscoring the complexity of the \gls{mv} segmentation task from echocardiography related to the inherent limitations of this imaging technique such as noise, artifacts and absence of clear intensity-based boundaries between intracardiac structures. However, the analysis also revealed a relatively good level of agreement among different operators, comparable to previous reported manual segmentation inter-uses variability \cite{Jassar_2011} and to the level of precision reached by the proposed automatic method. The process could be improved by manually annotating fewer annular points and exploiting an image registration approach to automate part of the process. In addition, the user may be asked to add an additional landmark on the aortic valve; this extra information could be used to automatically reorient the initial long-axis view to obtain a plane passing through the saddle horn and the mid-point of the posterior annulus. In this way, the end-user would segment the \gls{mv} starting always from the same cut-plane, thus making it easier to recognize anatomical landmarks and making the process more repeatable. Finally, the number of mutually parallel image planes that span the MV could be set automatically, for instance defining the shift between consecutive planes based on the voxel size. Again, also this modification could improve the repeatability of the process.\\
Even if the ensemble method adopted showed a significant improvement in the segmentation performance, the reported findings were limited to a small independent test set. The small dimension of the available dataset prevented also the application of the developed automated morphological analysis for a more comprehensive characterization of the \gls{mv} anatomies in patients affected by \gls{mr}, particularly in terms of variability in coaptation defects. The use of a larger dataset would enhance the pipeline's capability to efficiently and reliably analyze \gls{mr}-related anatomical characteristics of the \gls{mv}. Moreover, extending the method to a temporal segmentation problem could also help to increase its accuracy, as well as provide a comprehensive assessment of \gls{mv} anatomy over the entire cardiac cycle. In fact, some \gls{mv} substructures (e.g., the leaflets) are better imaged at specific phases of the cardiac cycle. Enforcing the anatomical continuity of segmented \gls{mv} structures in each frame during CNN training could enhance results and overcome limitations due to image quality. The inference time required by the proposed pipeline is still relatively high for real-time usage. The time efficiency can be improved by migrating all computational operations to GPU. Finally, the proposed pipeline can be used for diagnostics, patient selection and surgical treatment planning for transcatheter \gls{mv} repair. In the future, integration of the proposed pipeline into commercial ultrasound scanners could facilitate its validation in clinics and enable its use in cardiac interventions as a guidance during percutaneous repair of \gls{mv}.

\section{Conclusion}
\label{sec:conclusion}
A fully automated, 3D CNN-based pipeline was introduced for \gls{mv} segmentation from \gls{3dtee} and identification of anatomical landmarks and relevant features in regurgitant \gls{mv}s. The designed pipeline improves the workflow for \gls{mv} segmentation and anatomical quantification of the mitral annulus and leaflets, providing consistent and repeatable results in a comprehensive and fast way. This pipeline will potentially lead to a more reproducible and time-efficient quantification of \gls{mv} anatomy to support the diagnosis of \gls{mr} and planning of \gls{mv} repair, with a potential future extension to intraoperative support in transcatheter \gls{mv} repair surgeries.

\bibliographystyle{IEEEtran}
\bibliography{biblio}

\begin{IEEEbiography}[{\includegraphics[width=1in,height=1.25in, clip, keepaspectratio]{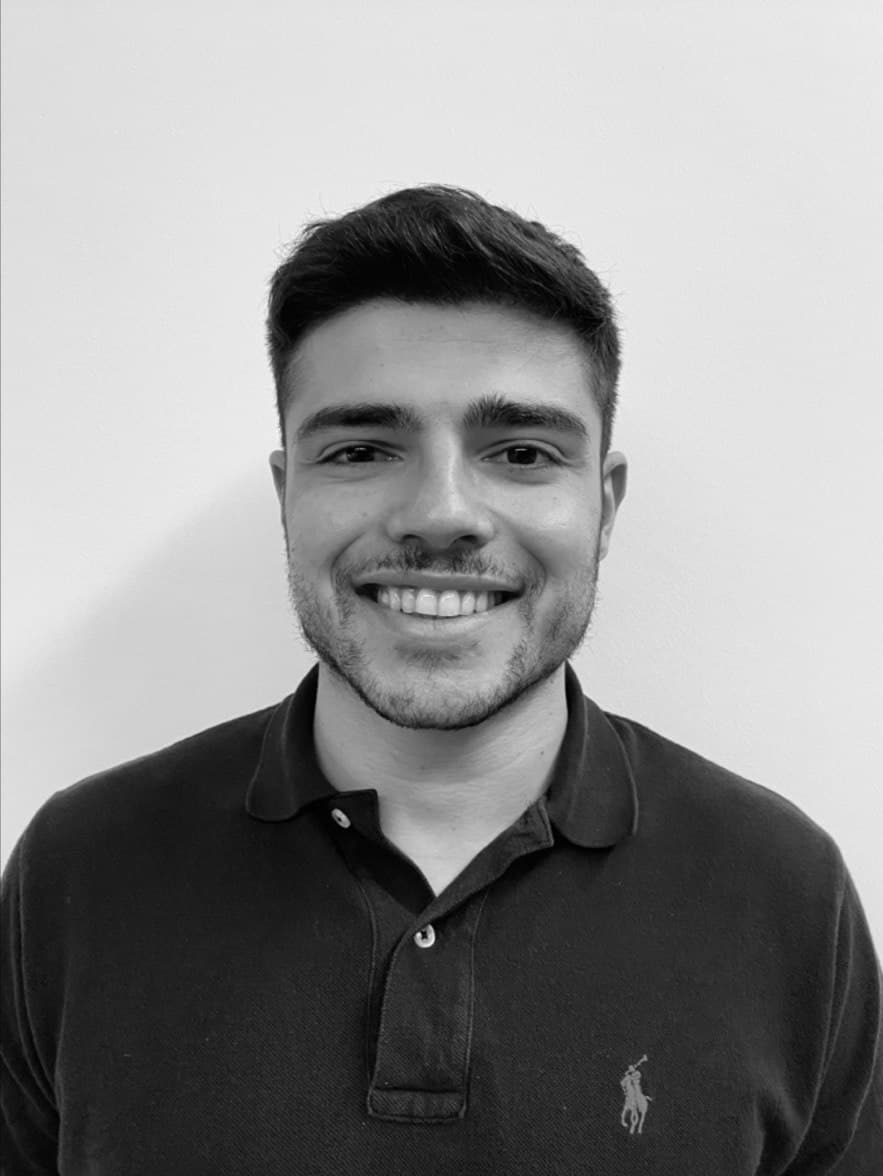}}]{Riccardo Munafò} received the M.S. degree in biomedical engineering from Politecnico di Milano, Milano, Italy, in 2021. He is currently enrolled as a PhD researcher in the Biomechanics Research Group from Politecnico di Milano; his research activity focuses on the development and application of AI-based methods for real-time segmentation of cardiac imaging and intracardiac catheter tracking. His research activity will contribute to developing the ARTERY EU-funded project.
\end{IEEEbiography}

\begin{IEEEbiography}[{\includegraphics[width=1in,height=1.25in, clip, keepaspectratio]{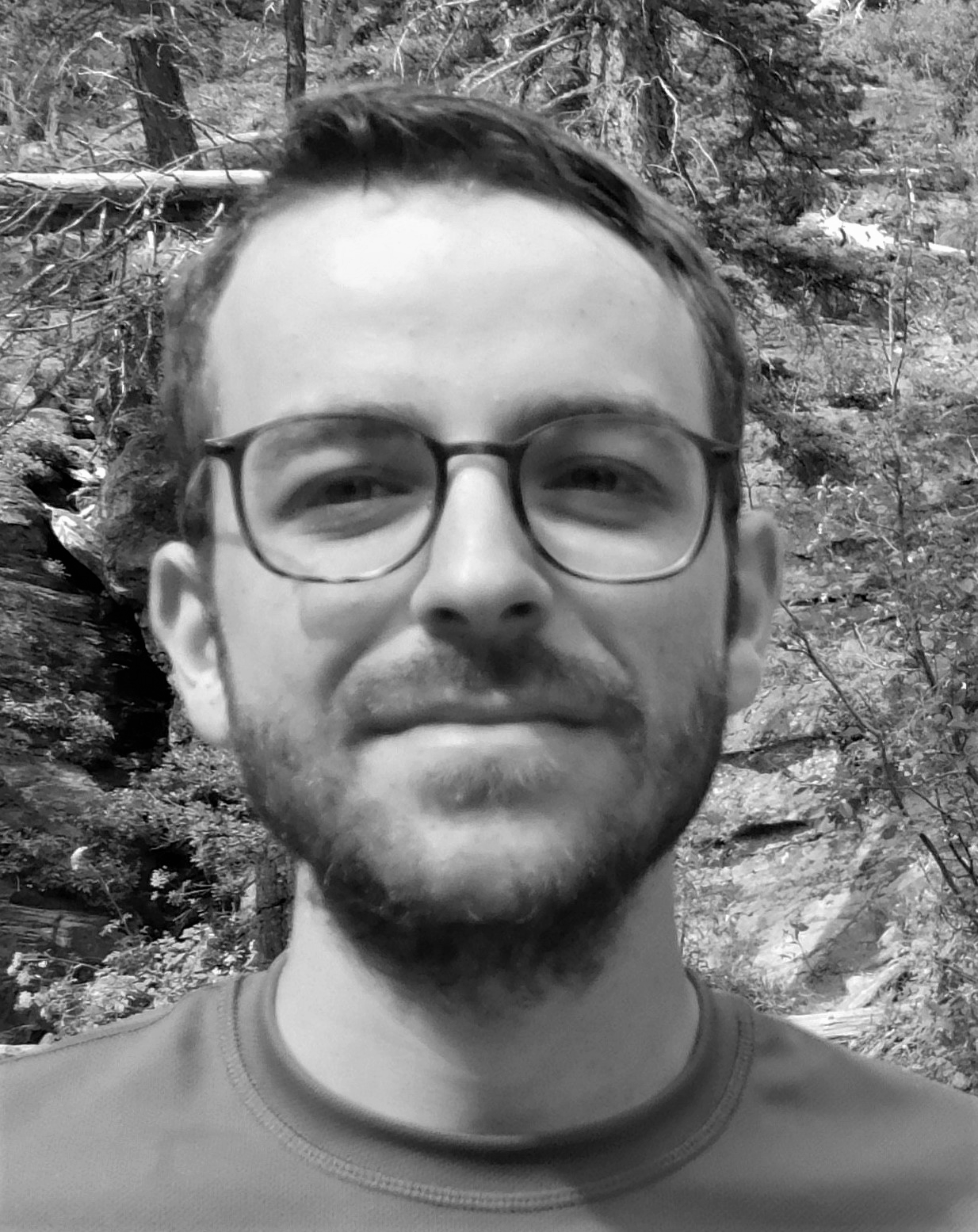}}]{Simone Saitta} received his Ph.D degree in Bioengineering from Politecnico di Milano in 2023. Over the years, he has worked as a researcher at Imperial College London, National University of Singapore and University of Cambridge. His research combines biomedical image analysis and simulation, focusing on the development of data-driven methods for clinical applications. Some of his research projects include 3D convolutional neural networks for medical image segmentation, implicit neural representations for image denoising and super-resolution, and automatic workflows for pre-procedural planning. He is currently a postdoctoral researcher in the Biomechanics Research Group at Politecnico di Milano. 
\end{IEEEbiography}

\begin{IEEEbiography}[{\includegraphics[width=1in,height=1.25in, clip, keepaspectratio]{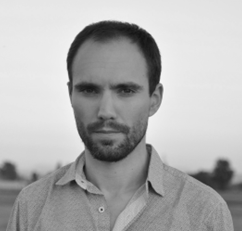}}]{Gicoamo Ingallina} received the MD Degree in 2011 from the Vita-Salute San Raffaele University and the Specialization in Cardiology from the University of Florence in 2017. He is currently Consultant Cardiologist in the Cardiovascular Imaging Unit at San Raffaele Hospital, Milan. His working activity focuses on the utilization of advanced cardiovascular imaging techniques for the comprehensive assessment and treatment of valvular heart diseases. His research is primarily devoted to the comprehensive utilization of echocardiography in valvular heart diseases and percutaneous interventions.
\end{IEEEbiography}

\begin{IEEEbiography}[{\includegraphics[width=1in,height=1.25in, clip, keepaspectratio]{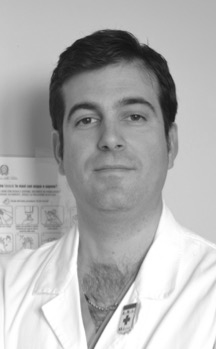}}]{Paolo Denti} Denti received the MD Degree and the Specialization in Cardiac Surgery from the Vita-Salute San Raffaele University, in 2004 and in 2009, respectively. He is currently responsible of transcatheter therapy as Hybrid Cardiovascular Surgeon at San Raffaele Hospital, Milan. His research is mainly in the field of percutaneous approach to structural heart disease (in particular mitral valve repair and aortic valve implantation), pre-clinical animal/bench study of new cardiac devices, minimally invasive and conventional heart surgery. His research activity will contribute to developing the ARTERY EU-funded project.
\end{IEEEbiography}

\begin{IEEEbiography}[{\includegraphics[width=1in,height=1.25in, clip, keepaspectratio]{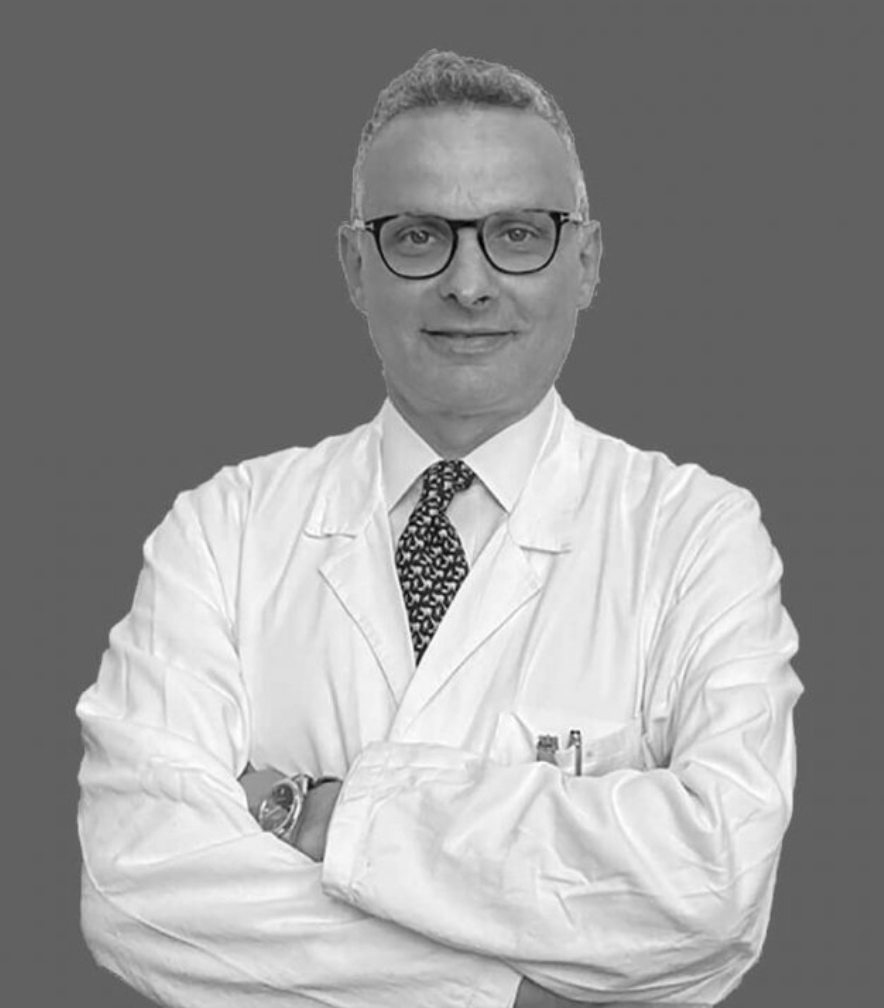}}]{Francesco Maisano} received MD Degree in 1990 from the Università Cattolica del Sacro Cuore and the Specialization in Cardiac Surgery from the Sapienza University of Rome in 1994. He is currently Chief of Cardiac Surgery Unit and Director of the Valve Center at San Raffaele Hospital, Milan. His research activities primarily focus on the percutaneous approach to structural heart disease. Over the years, he has collaborated with numerous partners to contribute to research and development of new devices and solutions. In recognition of his innovative work, he was awarded the Silver Medal for Innovation in 2018 by the European Society of Cardiology. He is currently Full Professor of Cardiac Surgery at Vita-Salute University. 
\end{IEEEbiography}

\begin{IEEEbiography}[{\includegraphics[width=1in,height=1.25in, clip, keepaspectratio]{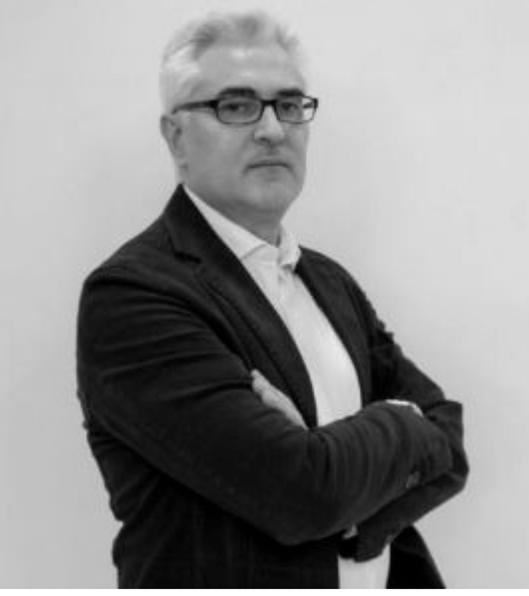}}]{Eustachio Agricola} received the MD Degree and the Specialization in Cardiology from the University of Siena, in 1996 and in 2000, respectively. He is currently Associate Professor of Cardiology of Vita-Salute University, and Head of Cardiovascular Imaging Unit, at San Raffaele Hospital, Milan. His research activity is mainly focused on the field of Cardiovascular Imaging, specializing in echocardiography. In recent years, he has devoted his research and clinical activities to the Interventional Echocardiography (echocardiographic support for percutaneous treatment of structural heart diseases). His expertise in valvular pathologies and commitment to exploring new percutaneous interventions make him a key figure in advancing the field.
\end{IEEEbiography}

\begin{IEEEbiography}[{\includegraphics[width=1in,height=1.25in, clip, keepaspectratio]{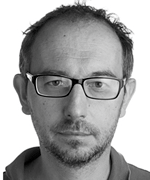}}]{Alberto Redaelli} received the M.S. degree in mechanical engineering and the Ph.D degree in bioengineering from Politecnico di Milano, Milano, Italy, in 1991 and in 1995, respectively. He is currently Full Professor of Biomechanics in the Department of Electronics, Information and Bioengineering of Politecnico di Milano and leads the Biomechanics reaserch group. Over his academic life, he has mainly focused on the development of new enabling tools, with the aim of finding new solutions to clinical problems. His research is mainly in the field of cardiovascular biomechanics and computational fluid dynamics, both experimental and computational, and embrace pioneering studies on fluid structure interaction approaches, design of innovative cardiovascular devices and prostheses, and valve mechanics studies with emphasis on patient specific modeling from imaging data. 
\end{IEEEbiography}

\begin{IEEEbiography}[{\includegraphics[width=1in,height=1.25in, clip, keepaspectratio]{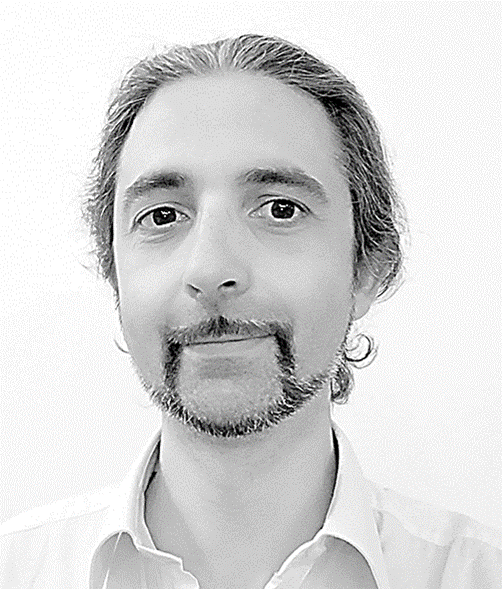}}]{Emiliano Votta} received the M.S. degree in biomedical engineering and the Ph.D degree in Bioengineering from Politecnico di Milano, Milano, Italy, in 2001 and in 2006, respectively. He is currently Associate Professor at the Department of Electronics, Information and Bioengineering, where he is part of the Biomechanics Research Group. He teaches two courses for Biomedical Engineering graduate students (Advanced Modeling Approaches For Cardiovascular Surgery [I.C.], and Computational Biomechanics Laboratory). His research activity is mainly devoted to computational modeling as a technology to support clinical decision making, mostly in the context of cardiovascular diseases and therapies. He is currently the coordinator of ARTERY, a European Project funded by the H2020 Programme that aims to revolutionize structural interventional cardiolody through robotics.

\end{IEEEbiography}

\EOD

\end{document}